\pdfoutput=1
\documentclass[11pt]{article}

\usepackage{amsmath}
\usepackage{amssymb}
\usepackage{array}
\usepackage{hyperref}
\usepackage[linesnumbered,ruled,vlined]{algorithm2e}
\usepackage{stfloats}  
\usepackage{algorithmic}
\usepackage{caption}
\usepackage{float}
\usepackage{pgfplots} 
\pgfplotsset{compat=1.18}
\usepackage{pgfplotstable}
\usepackage{tikz}
\usepackage{placeins}
\usepackage[preprint]{acl}

\usepackage{times}
\usepackage{latexsym}
\usepackage{booktabs}
\usepackage{adjustbox}
\usepackage{arydshln}
\usepackage{multicol}

\usepackage[T1]{fontenc}

\usepackage[utf8]{inputenc}

\usepackage{microtype}

\usepackage{inconsolata}

\usepackage{graphicx}
\usepackage{lipsum}
\usepackage{afterpage}

\title{TO-GATE: Clarifying Questions and Summarizing Responses with Trajectory Optimization for Eliciting Human Preference}

\author{
  Yulin Dou$^{\dagger}$ \quad Jiangming Liu$^{\dagger\ddagger*}$\\
  $^{\dagger}$School of Information Science and Engineering, Yunnan University, China \\
  $^{\ddagger}$Yunnan Key Laboratory of Intelligent Systems and Computing, China \\
  $^{*}$Corresponding Author: \texttt{jiangmingliu@ynu.edu.cn}
}

\begin{document}
\maketitle

\begin{abstract}
Large language models (LLMs) can effectively elicit human preferences through multi-turn dialogue. Complex tasks can be accomplished through iterative clarifying questions and final responses generated by an LLM acting as a questioner (STaR-GATE;~\citealt{andukuristar2024star}). However, existing approaches based on self-taught reasoning struggle to avoid irrelevant questions to the tasks. 
To address this limitation, we propose TO-GATE, a novel framework that enhances question generation through trajectory optimization, which consists of two key components: a clarification resolver that generates optimal questioning trajectories, and a summarizer that ensures task-aligned final responses. The trajectory optimization enables the model to produce effective elicitation questions and summary responses tailored to specific tasks.
Experimental results demonstrate that TO-GATE significantly outperforms baseline methods, achieving a 9.32\% improvement on standard preference elicitation tasks.

\end{abstract}

\section{Introduction}
The remarkable success of Large Language Models (LLMs) in various NLP tasks has given rise to a new paradigm of human-agent interaction, where users pose questions or instructions to LLM-based agents, which then generate responses \cite{mann2020language,brown2020language,reynolds2021prompt,madotto2021few,wang2023prompt,giray2023prompt,mayer2023prompt,barisin2024riesz,liu2024moe,yu2024multigprompt}. However, user queries can often be ambiguous due to implicit preferences~\cite{finn2018probabilistic,tamkintask2023task}. For instance, if a user requests a pasta recipe without specifying dietary restrictions (e.g., vegetarian), the agent may fail to provide a suitable response~\cite{andukuristar2024star}. To address this challenge, recent work has focused on preference elicitation and alignment techniques, ensuring that LLMs better adapt to complex human preferences and values \cite{li2025eliciting}.

\begin{figure}
    \centering
    \includegraphics[width=1.0\linewidth]{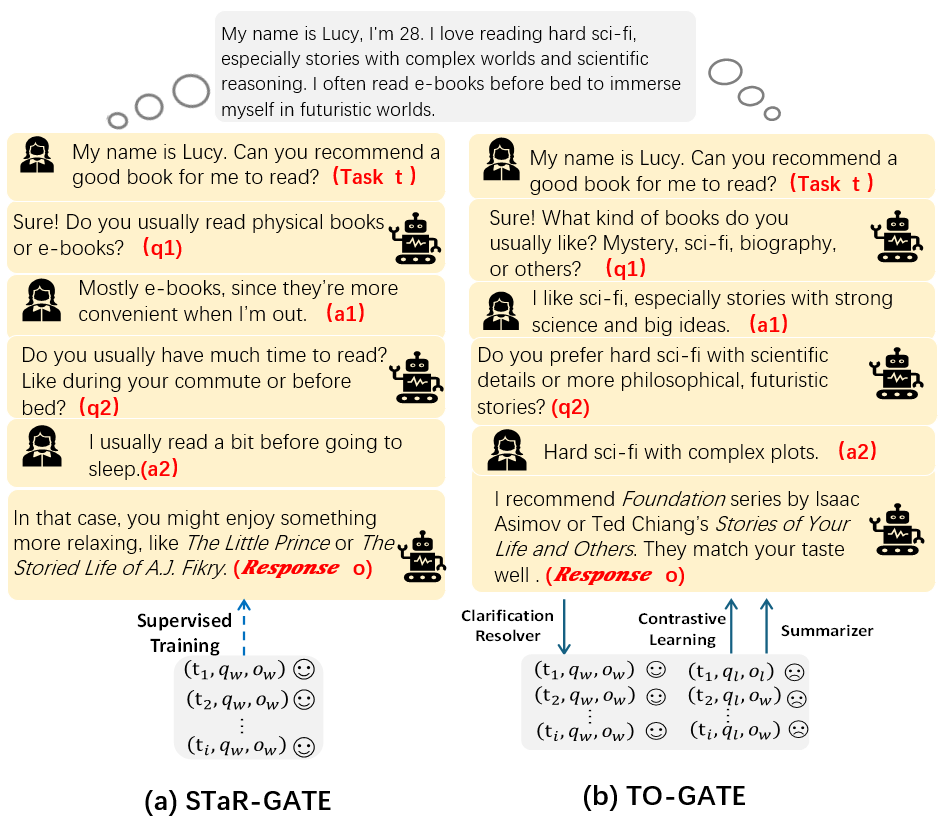}
    \caption{(a) STaR-GATE adopts the supervised training; (b) TO-GATE adopts contrastive learing with adoptive weights for find responses. TO-GATE’s response is better than STaR-GATE’s because TO-GATE recommends classic hard sci-fi that matches Lucy’s preference for complex worlds and scientific reasoning, while Star-GATE suggests lighter books that do not suit her interests.}
    \label{task}
\end{figure}

To enhance the ability of LLM-based agents to ask useful clarifying questions, STaR-GATE~\cite{andukuristar2024star} combines active preference elicitation (GATE;~\citealt{li2025eliciting}) with a self-improvement loop inspired by STaR~\cite{zelikman2022star}. During interactions, the agent iteratively refines its understanding by posing clarifying questions to elicit user preferences, ultimately generating a final response to the user original task via self-play, as illustrated in Figure~\ref{task}(a). However, STaR-GATE is trained solely on conversations with the highest conditional probability of gold responses, P(Gold Response|Conversations).
This optimization strategy fails to penalize degenerate dialogue trajectories, often resulting in ineffective question sequences that yield suboptimal task resolutions.

Motivated by the contrastive learning methods like Direct Preference Optimization (DPO;~\citealt{rafailov2023direct}), we propose TO-GATE that can clarify questions and summarize final responses with trajectory optimization, aiming to improve the ability of LLM-base agents to ask effective questions, which is shown in Figure~\ref{task}(b). The trajectory optimization consists of clarification resolver and summarizer. The clarification resolver adopts DPO strategies to positively reward effective conversations and penalize bad conversations, and summarizer adopts adaptive weights to enable questioner to summarize better responses based on the historical conversations. 



We evaluate our approach on standard preference elicitation tasks from GATE~\cite{li2025eliciting}, where model responses are concatenated and assessed by LLM judges. However, ~\cite{andukuristar2024star} identify significant position bias in the evaluations. To address this, we propose a deterministic evaluation protocol that averages scores across all possible response orderings.

Experimental results demonstrate that TO-GATE significantly outperforms existing baselines, establishing new state-of-the-art performance. Our analysis reveals that this improvement stems from two key factors:  effective question generation for preference elicitation, and high-quality final responses aligned with user intent.

The contributions of the paper are summarized:
\begin{itemize}
    \item We present TO-GATE, a novel training framework that jointly optimizes two critical capabilities: generating contextually-appropriate clarifying questions to resolve preference ambiguity, and producing accurate responses for task completion, aiming to specifically targets effective human preference elicitation.
    \item We extend the directed preference optimization for dynamical dialogue generation optimization, aiming to distinguish the good and poor conversations.
    \item We introduce a deterministic evaluation metric that eliminates position bias, providing stable and reproducible assessment scores. 
    \item  Experimental results demonstrate that our framework achieves SOTA performance on human preference elicitation benchmarks, outperforming existing approaches. The code and data will be released.
\end{itemize}



\section{Related Work}
\vspace{-0.5em}
\paragraph{Generative Active Task Eliciting}

Generative Active Task Elicitation leverages the interactive dialogue capabilities of language models to dynamically elicit user preferences, offering a novel paradigm for resolving task ambiguity~\cite{li2025eliciting,piriyakulkij2023active,lin2024decision,franken2023social,handa2024bayesian,aliannejadi2021building}. The framework of generative active task elicitation (GATE) positions language models as \textit{active questioners}, breaking away from the traditional reliance on static prompts~\cite{brown2020language} that explicitly ask users to declare their preferences. GATE address the task ambiguity through multi-turns dialogue, where the language models autonomously generate a sequence of questions designed to maximize informational value. Similarly, \citet{hong2023zero} explore the feasibility of transferring large model guidance capabilities to lightweight models. They utilize GPT-3.5 to simulate human-machine interactions, incorporating constitutional AI agents~\cite{bai2022constitutional} to correct generated content and providing a solution for resource-constrained environments.
\vspace{-0.5em}
\paragraph{Self-Taught Reasoner}
Self-taught reasoning enables language models to autonomously enhance their reasoning abilities by generating and utilizing intermediate questions and answers. Self-Taught Reasoner (STaR;~\citealt{zelikman2022star}) demonstrates significant performance improvements in arithmetic and symbolic reasoning tasks, which constructs a training loop where the model first generates synthetic reasoning traces (such as chain-of-thought) and then fine-tunes on these self-generated data. Several variants of STaR have been proposed to further improve the reasoning ability of LLMs. V-STaR (\citealt{hosseini2024v}) shows that training a verifier to guide reasoning generation also significantly improves performance. Quiet-STaR (\citealt{zelikman2024quiet}) focuses on generating more concise and effective reasoning paths, aiming to guide the model to output the minimal yet crucial reasoning steps, thereby reducing redundant information. 
\vspace{-0.5em}
\paragraph{Direct Preference Optimization}
Direct Preference Optimization (DPO;~\citealt{rafailov2023direct}) maximizes the likelihood of preferred responses in human preference data, avoiding the explicit construction of a reward model and the instable training of traditional reinforcement learning. However, the standard DPO motivated by the contrastive learning, primarily focuses on pairs of instances, which limits its effectiveness in multi-turn dialogues. To address this limitation, several DPO-based methods for multi-turn alignment have been proposed. Extended Turn-level Optimization (ETO;~\citealt{song2024trial}) extends the DPO loss function to each turn in multi-turn dialogues, aiming to achieve multi-turn alignment. However, this approach has limitations in terms of alignment granularity and theoretical guarantees. Direct Multi-turn Preference Optimization (DMPO;~\citealt{shi2024direct}) introduces a State-Action Occupancy Measure (SAOM) constraint and applies length normalization to the Bradley-Terry model, theoretically eliminating the partition function $Z$. Segment-Level Direct Preference Optimization (SDPO;~\citealt{kong2025sdpo}) further refines the alignment granularity by dynamically selecting key segments within dialogues for optimization. In this work, we combines self-learning reasoning techniques with the model's inherent reasoning capabilities to guide user preferences, thus enhancing interaction quality in multi-turn dialogues.

\section{Problem Definition}
According to the previous works~\cite{andukuristar2024star,li2025eliciting}, we consider the problem of eliciting human preference as learning an effective questioning policy for personalized generation tasks. The set of tasks is defined as 
\[\mathcal{T} = \{t_1, t_2, \dots, t_{|\mathcal{T}|}\}\]
where each task \(t_i\) represents a specific generation task associated with a specific persona from the user persona set:  
\[\mathcal{U} = \{u_1, u_2, \dots, u_{|\mathcal{U}|}\}\]  
where each \(u_j\) represents the personalized information about a user, such as their goals, tone, preferences, and background knowledge.

Effective personalized generation tasks require three specialized models.
\begin{itemize}
    \item Oracle, \(O\), that is allowed to access to both the task \(t_i\) and the user persona \(u_j\) can generate a gold response \(o^g_{{ij}} \sim p_O(o^g_{ij} \mid t_i, u_j)\), serving as the target for the personalized generation task.
    \item Roleplayer, \(R\), that is also allowed to access to both the task and persona information, simulates user behaviors and responds to the questions posed by the Questioner.
    \item Questioner, \(Q\), that is the  model we need to optimize, can access the task \(t_i\), but not the user persona \(u_j\). Questioner is trained to obtain the questioning policy $\pi$ that is used to ask effective questions for eliciting human preference.
\end{itemize}
Questioner engages in multi-turn dialogue with Roleplayer to progressively infer latent user persona attributes through iterative preference elicitation. This dynamic interaction enables the generation of responses that are progressively personalized to the emergent user profile.

The optimized models utilize gold responses provided by Oracle as supervision signals, while the complete conversational trajectory and true user persona \(u_j\) remain unobserved by Questioner. Formally, we optimize the questioning strategy to maximize the log-likelihood of generating personalized responses \(s_{ij}\) that match Oracle's gold response \(o^g_{ij}\):
\begin{equation}
\begin{aligned}
J(Q, R, & \mathcal{T}, \mathcal{U}) = \\
&\sum_{t_i \in \mathcal{T}} \sum_{u_j \in \mathcal{U}} 
\mathbb{E}_{s_{ij}} \left[\log p_{Q_{\text{BASE}}}(o^g_{ij} \mid t_i, s_{ij})\right]
\end{aligned},
\end{equation}
where, \(s_{ij} = [q_{ij1}, a_{ij1}, \dots, q_{ijk}, a_{ijk}]\) is a conversation between Questioner and Roleplayer in \(K\)-turns conversation.
At each turn, Questioner generates a question \(q_{ijk}\) based on the task and the conversation history:
\[q_{ijk} \sim p_Q(q \mid t_i, q_{ij1}, a_{ij1}, \dots, q_{ij(k-1)}, a_{ij(k-1)})\]
Role-player generates a answer \(a_{ijk}\) based on the task and user persona:
\[a_{ijk} \sim p_R(h \mid u_j, t_i, q_{ij1}, a_{ij1}, \dots, q_{ijk})\]

In the problem definition, $p_O$ and $p_R$ is fixed because Oracle and Role-player are assumed that they know everything about the users, while $p_Q$ need to be optimized by adjusting the questioning policy $\pi$, which designs informative multi-turn questions for eliciting useful information from the user. Specifically, $p_{Q_{\text{BASE}}}$ refers to a frozen baseline model used for evaluation. It measures the likelihood of generating the gold response $o^g_{ij}$ given the task $t_i$ and the conversation sequence $s_{ij}$.

\section{TO-GATE}
\begin{figure*}[t]
    \centering
    \includegraphics[width=0.9\linewidth]{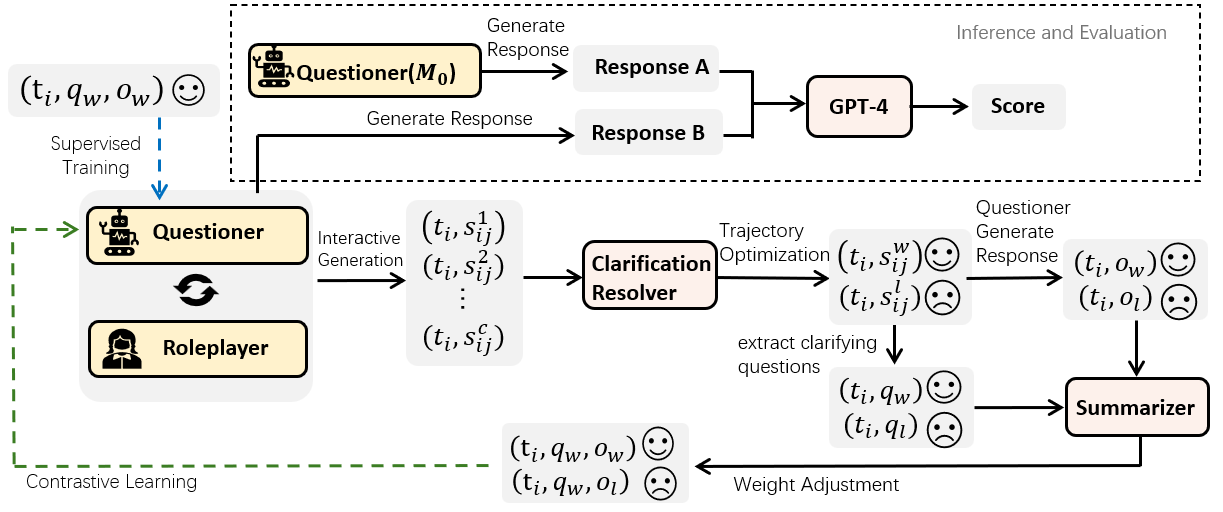}
    \caption{In the training phase, TO-GATE optimizes the clarification resolver by using contrastive learning to distinguish between positive and negative conversations; at the same time, it enhances the summarizer’s performance by adjusting loss weights to differentiate between questions and final responses. In the inference and evaluation phase, GPT-4 is employed to judge the simulated dialogue responses generated by the base model $M_0$ and the trained model $M_n$.
    }
    \label{to-gate}
\end{figure*}

As shown in Figure~\ref{to-gate}, we propose the framework of TO-GATE for eliciting human preference with trajectory optimization. TO-GATE starts by training an initial agent through supervised fine-tuning, making it have basic elicitation ability.\footnote{The initialization phrase is similar to STaR-GATE}
Questioner interacts with Roleplayer to explore dialogue trajectories in a trial-and-error fashion. Through iterative refinement, it progressively improves the clarification resolver, thereby enhancing the model’s ability to elicit and capture user preferences effectively. Finally, the system summarizes the dialogue history to generate a personalized response tailored to the original task and the specified persona.

\subsection{Clarification Resolver}

Clarification resolver adopts Direct Preference Optimization (DPO;~\citealt{rafailov2023direct}) to optimize the policy model's preference for high-quality responses by leveraging human preference data through contrastive learning. 
\vspace{-0.5em}
\paragraph{DPO} 
Traditional Direct Preference Optimization methods typically train on a static set of human preference data. They optimize the model to favor generating high-quality responses that are preferred by humans by comparing preferred answers with less preferred ones.
The core idea of this approach is “preference contrastive learning,” where the model learns to prefer the better answer over the worse one given the same input.
In traditional DPO, for each user input $x$, given a preferred response $y_w$ (positives) and a less preferred response $y_l$ (negatives), the DPO objective is defined as:
\begin{align}
\label{eq:dpo-loss}
L_{\text{DPO}} & (\pi_{\theta}; \pi_{\text{ref}}) = - \mathbb{E}_{(x, y_w, y_l) \sim D} \log \sigma \notag \\
&\left[ \beta \log \frac{\pi_{\theta}(y_w \mid x)}{\pi_{\text{ref}}(y_w \mid x)} 
- \beta \log \frac{\pi_{\theta}(y_l \mid x)}{\pi_{\text{ref}}(y_l \mid x)} 
\right],
\end{align}
where, $\pi_{\theta}$ denotes the trainable policy model, $\pi_{\text{ref}}$ represents the reference policy (typically a pretrained base model), $\beta$ is a temperature parameter that controls the sensitivity to preference differences, and $\sigma$ is the sigmoid function. This objective essentially quantifies how much more the policy $\pi_{\theta}$ prefers the human-selected output $y_w$ over the rejected output $y_l$, relative to the reference model. 

As such, DPO is essentially a contrastive loss function that relies entirely on static preference data and does not involve interactive dynamic updates. This makes it difficult to effectively capture contextual dependencies in multi-turn dialogues and limits its ability to dynamically explore and discover better clarification strategies. To overcome these limitations, we introduce a multi-turn trajectory exploration optimization strategy based on the DPO loss.
\vspace{-0.5em}
\paragraph{Trajectory Exploration}
Trajectory exploration follows an iterative loop of ``exploration-collection-training,'' where new positive and negative data are actively generated and filtered to enrich the training set. The objective of this exploration-based strategy follows the reinforcement learning paradigm and is formulated as:
\begin{equation}
\begin{aligned}
\max _{\pi_\theta} & \mathbb{E}_{(t_i,u_j) \sim \mathcal{D}, q_{ij} \sim \pi_\theta(t_i \mid q_{ij})}[r(t_i, q_{ij})] - \\
& \beta \mathbb{D}_{\text{KL}}\left[\pi_\theta(q_{ij} \mid t_i) \, \|\, \pi_{\text{ref}}(q_{ij} \mid t_i)\right],
\label{eq:objective}
\end{aligned}
\end{equation}
where, $\pi_{\theta}$ and $\pi_{\text{ref}} $ denotes the policy model (e.g., Questioner), and $q_{ij}$ represents a clarification-oriented multi-turn questions associated with the task. The reward function $r(t_i, q_{ij})$ is designed to evaluate the quality of the multi-turn questions with respect to both the original task and the clarification context.
The KL divergence term, weighted by the parameter $\beta$, serves as a regularization mechanism that constrains the policy deviation from the reference model, ensuring that the optimization process maintains a balance between preference-driven flexibility and training stability. 

\begin{figure}
    \centering
    \includegraphics[width=0.95\linewidth]{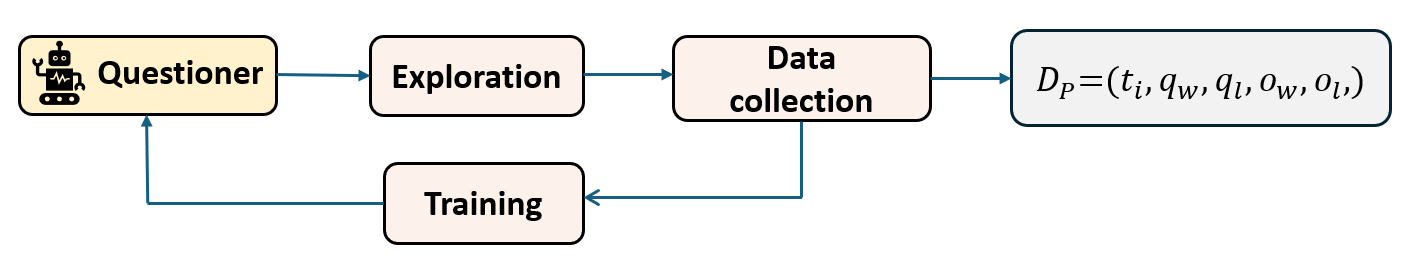}
    \caption{The dynamic dataset $D_p$ is generated and updated through a continuous cycle of ``exploration-collection-training,'' enabling dynamic expansion of the training data and continuous improvement of model performance.}
    \label{fig:dp}
\end{figure}

\vspace{-0.5em}
\paragraph{Optimization}
The positive-negative trajectory pair $(t_i, q_w, q_l)$ can be modeled using the Bradley-Terry (BT;~\citealt{bradley1952rank}) to obtain the probability distribution of human preferences p as follows:

\begin{equation} 
    \begin{aligned}
        p\left(q_{w} \succ q_{l} \mid t_i\right)=
        \frac{\exp \left(r\left(t_i, q_{w}\right)\right)}{\exp \left(r\left(t_i, q_{w}\right)\right)+\exp \left(r\left(t_i, q_{l}\right)\right)}
    \label{eq:BT}
    \end{aligned}.
\end{equation}
Based on the optimal policy defined in Eq.~\eqref{eq:objective}, the reward function can be expressed as:
\begin{equation} 
r(t_i, q_{ij})=\beta \log \frac{\pi_{r}(q_{ij} \mid t_i)}{\pi_{\mathrm{ref}}(q_{ij} \mid t_i)}+\beta \log Z(t_i),
\label{eq:reward function}
\end{equation}
where $Z(t_i)$ is the partition function:
\begin{equation}
Z(t_i) =  {\textstyle \sum_{q_{ij}}}  \pi_{\text{ref}}(q_{ij} | t_i) \exp \left( \frac{1}{\beta} r(t_i, q_{ij}) \right).
\end{equation}
Substitute Eq.~\eqref{eq:reward function}, into Eq.~\eqref{eq:BT} to get the BT model over policy:
\begin{equation}
\begin{aligned}
&p\left (q_{w} \succ q_{l} \mid t_i\right) = \\
&\sigma\left(\beta \log \frac{\pi_{\theta}\left(q_{w} \mid t_i\right)}{\pi_{\theta}\left(q_{l} \mid t_i\right)}-\beta \log \frac{\pi_{\mathrm{ref}}\left(q_{w} \mid t_i\right)}{\pi_{\mathrm{ref}}\left(q_{l} \mid t_i\right)}\right),
\end{aligned}
\end{equation}
where $\sigma$ is the sigmoid function. Then the optimal policy $\pi_\theta$ of TO-GATE can be obtained by applying the DPO method in Eq.~\eqref{eq:dpo-loss}.
\begin{equation}
\begin{aligned}
&{L}_{c} = -\mathbb{E}_{(t_i, q_w, q_l) \sim \mathcal{D}_p} \\ 
&\left[ \log \sigma \Big( \beta \log \frac{\pi_{\theta}(q_w | t_i)}{\pi_{\theta}(q_l | t_i)} - \beta \log \frac{\pi_{\text{ref}}(q_w | t_i)}{\pi_{\text{ref}}(q_l | t_i)} \Big) \right].
\label{eq:Trajectory dpo loss}
\end{aligned}
\end{equation}
where, $D_p = {(t_i, q_w, q_l)}$ denotes the dynamic training dataset generated in each iteration through the interaction between the Questioner and the Roleplayer, facilitated by the Clarification Resolver.
Each sample in $D_p$ consists of the task input $t_i$, a preferred clarification $q_w$ (positive clarification), and a less preferred clarification $q_l$ (negative clarification), both generated by the model. Unlike a static dataset, $D_p$ is constructed and continuously expanded through an iterative loop of exploration, data collection, and training, which is shown in Figure~\ref{fig:dp}.


\subsection{Summarizer}
The ultimate goal of eliciting human preference is to generate high-quality and personalized final responses. Questioner needs to raise questions and summarize the history to give the final response as well. To better align the training objective with this end task, we propose a summarizer that differentiates training losses between the multi-turn clarification dialogues and the final response.


We categorize the Questioner's generations into two distinct types based on their functional role in preference elicitation: 
\begin{itemize}
    \item \textbf{Clarifications} are multi-turn questions collected via trajectory exploration, $(q_{ij1}, q_{ij2}, .. q_{ij(K-1)})$.
    \item \textbf{Responses} are generated based on clarifications by prompting the model to produce final responses, i.e., $q_{ijK}$, which we denote as $o_{ij}$ for simplicity.
\end{itemize}
Accordingly, the DPO loss function for the final response stage is defined as follows:
\begin{equation}
\begin{aligned}
&{L}_{o} = -\mathbb{E}_{(t_i, o_w, o_l) \sim \mathcal{D}_p} \\ 
&\left[ \log \sigma \Big( \beta \log \frac{\pi_{\theta}(o_w | t_i)}{\pi_{\theta}(o_l | t_i)} - \beta \log \frac{\pi_{\text{ref}}(o_w | t_i)}{\pi_{\text{ref}}(o_l | t_i)} \Big) \right].
\label{eq:response dpo loss}
\end{aligned}
\end{equation}
${L}_o$ measures the model’s ability to fit user preference signals in the final output. $o_w$ (positive final response) represents the preferred final response. $o_l$ (negative final response) represents the less preferred final response.

To differentiate the contribution of each part, we introduce separate loss weights for the two stages. Specifically, the DPO loss is computed as a weighted sum of the losses over clarifications and responses:

\begin{equation}
\begin{aligned}
{L} = \frac{1}{1 + \lambda} \ \cdot L_c +  \frac{\lambda}{1 + \lambda} \cdot L_o,
\end{aligned}
\label{eq:weight}
\end{equation}
where ${L}_c$ and ${L_o}$ correspond to looses on clarifications and responses, respectively. The hyperparameter $\lambda $ is used to balance clarifications and responses.

\subsection{Training}
The core objective of this optimization is to increase the probability of successful trajectories ${q_w}$ and decrease the probability of failed trajectories ${q_l}$. 
The trajectory exploration is designed to jointly improve both clarification and summarizer capabilities of the model.
The training process of TO-GATE is shown in Algorithm ~\ref{alg:to-star_algorithm}.

\begin{algorithm*}[t]
\caption{TO-GATE}
\label{alg:to-star_algorithm}
\KwIn{Initial policy $\pi_{\text{0}}$, Task set $T = \{t_i\}_{i=1}^D$, Persona set $U = \{u_j\}_{j=1}^C$, Gold responses $G = \{o^g_{ij}\}, i\in[1,D], j\in[1,C]$}
\KwOut{Final policy $\pi_\theta$}

\Repeat{ max iterations}{
    \tcp{Trajectory Optimization Phase}
    \ForEach{$t_i \in T, u_j \in U$}{
        Generate conversation samples: $\{s_{ij}^c\}_{c=1}^{10} \leftarrow \pi_{0}(t_i, u_j)$\;
        Select best and worst conversations by base policy likelihood:
            $s_{ij}^w = \arg\max_{s_{ij}^c} \log p_{\pi_{\text{0}}}(o^g_{ij} | t_i, s_{ij}^c), \quad
            s_{ij}^l = \arg\min_{s_{ij}^c} \log p_{\pi_{\text{0}}}(o^g_{ij} | t_i, s_{ij}^c)$
        
        Generate responses:
            $o_w = \pi_{0}(t_i, s_{ij}^w), \quad
            o_l = \pi_{0}(t_i, s_{ij}^l)$
        
        extract clarifying questions:
            $q_w \leftarrow \mathrm {extract}(s_{ij}^w, t_i), \quad
            q_l \leftarrow \mathrm{extract}(s_{ij}^l, t_i) $
    }

    \tcp{Training Phase}
    Optimize $\pi_0$ via supervised fine-tuning: 
    $\mathcal{L}_{\mathrm{SFT}}(\pi_0) = -\mathbb{E}_{(t_i, q_w, o_w) \sim D} \left[ \log \pi_0\left((q_w, o_w) \mid t_i\right) \right] \quad \pi_1 \gets \pi_0
    $\\

    Update reference policy: \quad
        $\pi_{\mathrm{ref}} \gets \pi_1$
        
    Execute Trajectory Optimization Phase again to build contrastive dataset:
     $D_p = (t_i, q_w, q_l, o_w, o_l)^{(i)}$

    Optimize $\pi_1$ according to Eq.~(\ref{eq:Trajectory dpo loss})~(\ref{eq:response dpo loss})~and~(\ref{eq:weight})   
    \quad $\pi_2 \gets \pi_1$

}
\Return $\pi_\theta$;
\end{algorithm*}

\subsection{Evaluation}

Our objective is obtaining the well-trained Questioner that can ask clarifying questions to elicit human preference, and generate the correct responses to the tasks by summarizing the historical conversations. The evaluation is required to evaluate the clarifications and responses.

\paragraph{Clarification Metric}
To measure the clarification of the trained model $Q$, We use the log-probability of gold responses $o^g_{ij}$ conditioning on the simulated dialogue history $s_{ij}$ given by $Q$:
\begin{equation}
\log p_{Q_{\text{BASE}}}(o^g_{ij} \mid t_i, s_{ij}),
\end{equation}
where $Q_{\text{BASE}}$ is a base language model without training. High log-probability means that the simulated dialogues have high probability to generate the gold response, showing that the well-trained $Q$ can generate clarifying questions.
\vspace{-0.5em}
\paragraph{Response Metric}
We use \textit{win rate} to evaluate response quality by comparing the responses from the trained $Q$ to the responses from the base model $Q_{\text{BASE}}$. The responses of the two models are concatenated and fed to GPT-4 for judgments. Aiming to avoid the bias of the concatenation order, we propose the deterministic dual-pass evaluation metric. The responses of the trained $Q$ precedes the responses of the base model for the first pass evaluation, and reversing them for the second pass evaluation.

\section{Experiments}
According to the previous work~\cite{li2025eliciting}, Experiments are carried out on the tasks of eliciting human preference to validate the effectiveness of the proposed methods. 

\subsection{Dataset}
We use a subset of human queries from the open-source Instruct Human-Assistant Prompt Dataset\footnote{\url{https://huggingface.co/datasets/Dahoas/instruct-human-assistant-prompt/}} to construct the initial task set and obtain high-quality persona set from the PRODIGY \cite{occhipinti2023prodigy}.
We enumerate pairs of task and user persona $(t_i, u_j)$ to generate corresponding gold response $o^g_{ij}$ by applying GPT-4 as Oracle.\footnote{Details can be found in Appendix~\ref{sec:dataset-construction}}






\subsection{Models}
To evaluate the effectiveness of our proposed TO-GATE training framework, we compare our model against the following representative baselines:
\begin{itemize}
    \vspace{-0.5em}
    \item \textbf{STaR-GATE}~\cite{andukuristar2024star}~~~~The model fine-tuned via supervised learning on positive trajectories only.
    \vspace{-0.5em}
    \item \textbf{DPO}~~~~The model trained using Direct Preference Optimization without any prior supervised fine-tuning.
    \vspace{-0.5em}
    \item \textbf{TO-GATE}~~~~Our model trained using the trajectory optimization.
\end{itemize}
According the previous work~\cite{andukuristar2024star}, we denote the corresponding model as $M_n$ if they are trained in $n$ epoch. All the models are trained in three epochs and we choose the best one as the final model.

\subsection{Training and Evaluation Settings}
We use Mixtral-7B-Instruct-v0.2 as Questioner and Mixtral-8x7B-Instruct as Roleplayer.
In supervised fine-tuning, we set the batch size to 4, employ a learning rate of $2.0 \times 10^{-5}$, apply a 10\% warm-up ratio, and use a cosine learning rate scheduler.  In direct preference optimization, we set batch size to 4 and the learning rate to $1.0 \times 10^{-6}$. The preference strength parameter $\beta$ in the DPO loss is fixed at 0.1. The $\lambda$ is set to 0.33. The training are conducted on two NVIDIA A100 GPUs with 80GB.
Aiming to evaluate the performances of the model on eliciting human preference, we check the win rate by comparing all the models to $M_0$ that is without any fine-tuning.

\subsection{Results}

\vspace{-0.5em}
\paragraph{Response Evaluation}
Table~\ref{tab:win_rates} shows results of responses across models comparing to $M_0$. The proposed TO-GATE achieves the highest win rates across all comparative experiments, demonstrating that the generated responses effectively align with user personas by capturing human preferences. In contrast, traditional DPO models, which are trained on both positive and negative examples, tend to underperform compared to supervised models like STaR-GATE, which rely solely on positive examples. This performance gap arises because DPO is more susceptible to noise introduced by the automatically generated trajectories from the Questioner and Roleplayer modules without dynamic training data (shown in Eq.~\ref{eq:Trajectory dpo loss}). In response, our TO-GATE model leverages supervised training to initialize the Questioner, thereby minimizing the impact of noisy trajectory generation and leading to superior performance. 

\begin{table}[!tp]
\centering
\begin{tabular}{lccc}
\toprule
\textbf{Models} & \textbf{A-B} & \textbf{B-A} & \textbf{Average} \\
\midrule
STaR-GATE & 82.00 & 65.67 & 73.83 \\
DPO & 73.66 & 49.00 & 61.33 \\
\hdashline
TO-GATE & \textbf{89.90} & \textbf{76.33} & \textbf{83.15} \\
\bottomrule
\end{tabular}
\caption{Results of responses across models, where A-B and B-A means that the first and second part of the dual-pass evaluation, respectively. The results of dual-pass evaluation is averaged to be the final score.}
\label{tab:win_rates}
\end{table}

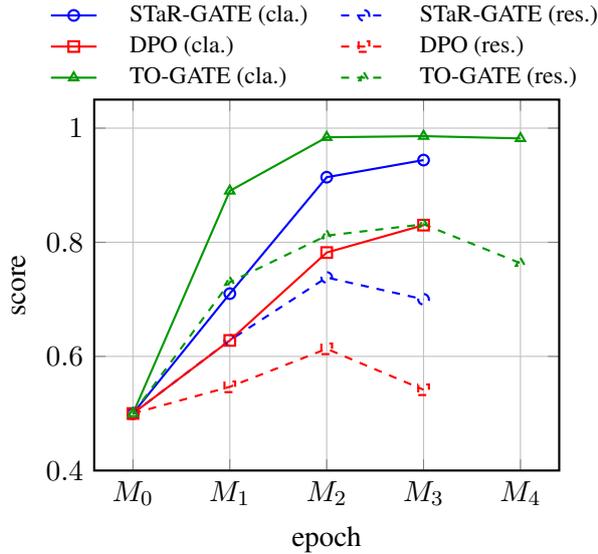
\begin{figure}[!tp]
    \centering
    \begin{tikzpicture}
        \begin{axis}[
            width=\linewidth,
            height=6.5cm,
            xlabel={epoch},
            ylabel={score},
            xtick={0,1,2,3, 4},
            xticklabels={$M_0$, $M_1$, $M_2$, $M_3$, $M_4$},
            ymin=0.4, ymax=1.05,
            grid=both,
            legend style={
                at={(0.5,1)},
                anchor=south,
                draw=none,
                fill=none,
                font=\small,
                legend columns=2,
                column sep=10pt
            },
            legend cell align={left},
            thick
        ]

        \addplot[color=blue, mark=o, solid] coordinates {(0, 0.5) (1, 0.710) (2, 0.914) (3, 0.944)};
        \addlegendentry{STaR-GATE (cla.)}

         \addplot[color=blue, mark=o, dashed] coordinates {(0, 0.5) (1, 0.6283) (2, 0.7383) (3, 0.7000)};
        \addlegendentry{STaR-GATE (res.)}

        \addplot[color=red, mark=square, solid] coordinates {(0, 0.5) (1, 0.628) (2, 0.782) (3, 0.830)};
        \addlegendentry{DPO (cla.)}

        \addplot[color=red, mark=square, dashed] coordinates {(0, 0.5) (1, 0.5467) (2, 0.6133) (3, 0.5416)};
        \addlegendentry{DPO (res.)}
        \addplot[color=green!60!black, mark=triangle, solid] coordinates {(0, 0.5) (1, 0.890) (2, 0.984) (3, 0.986) (4, 0.982)};
        \addlegendentry{TO-GATE (cla.)}

        \addplot[color=green!60!black, mark=triangle, dashed] coordinates {(0, 0.5) (1, 0.7300) (2, 0.8117) (3, 0.8315) (4, 0.7633)};
        \addlegendentry{TO-GATE (res.)}
        






        \end{axis}
    \end{tikzpicture}
    \caption{Results of clarification and responses across models in different epochs. Solid curves indicate clarification scores, and dashed curves indicate response scores.}
    \label{fig:combined-win-rates}
\end{figure}

\vspace{-0.5em}
\paragraph{Clarification Evaluation}
The solid curves in Figure~\ref{fig:combined-win-rates} illustrate the clarification performance across different models. As training progresses, the log-probabilities of gold responses given the simulated dialogues increase consistently (i.e., $M_0 < M_1 < M_2 < M_3$), indicating steady improvements in the models' ability to generate effective clarifying questions. TO-GATE significantly outperforms both STaR and DPO across all iterations, demonstrating superior clarification capabilities.
\vspace{-0.5em}
\paragraph{Response Versus Clarification}
The dashed curves in Figure~\ref{fig:combined-win-rates} shows the response performance across different models. When considered alongside the solid curves, it is evident that more effective clarifying questions generally lead to improved responses. However, after two training epochs (i.e., M3), a notable divergence emerges between the trends for clarifications and responses. This discrepancy suggests that while the ability to generate effective clarifying questions continues to improve, it does not always result in a consistent, monotonic enhancement in the quality of  responses. Despite this, our model still benefits from improved clarifications, as we explicitly separate the processes of clarification and response generation by introducing a dedicated response loss, as detailed in Section 4.2. In addition, our model can be further improved if having more training epochs according to the tend of reults curves.


\begin{figure}[ht]
    \centering
    \begin{tikzpicture}
        \begin{axis}[
            width=\linewidth,
            height=6.5cm,
            xlabel={epoch},
            ylabel={score},
            xtick={0,1,2,3},
            xticklabels={$M_0$, $M_1$, $M_2$, $M_3$},
            ymin=0.4, ymax=1.05,
            grid=both,
            legend style={
                at={(0.5,0.7)},
                anchor=south,
                draw=none,
                fill=none,
                font=\small,
                legend columns=1,
                column sep=10pt
            },
            legend cell align={left},
            thick
        ]

        \addplot[color=purple, mark=star] coordinates {(0, 0.5) (1, 0.7300) (2, 0.8117) (3, 0.8315)};
        \addlegendentry{TO-GATE}

        \addplot[color=orange, mark=diamond] coordinates {(0, 0.5) (1, 0.6616) (2, 0.7300) (3, 0.7750)};
        \addlegendentry{TO-GATE w/o clarification resolver}
        
        \addplot[color=green!60!black, mark=triangle] coordinates {(0, 0.5) (1, 0.7150) (2, 0.8283) (3, 0.8100)};
        \addlegendentry{TO-GATE w/o summarizer}
        




        \end{axis}
    \end{tikzpicture}
    \caption{Results of responses given by models with ablations in different epochs.}
    \label{fig:W/o weight}
\end{figure}
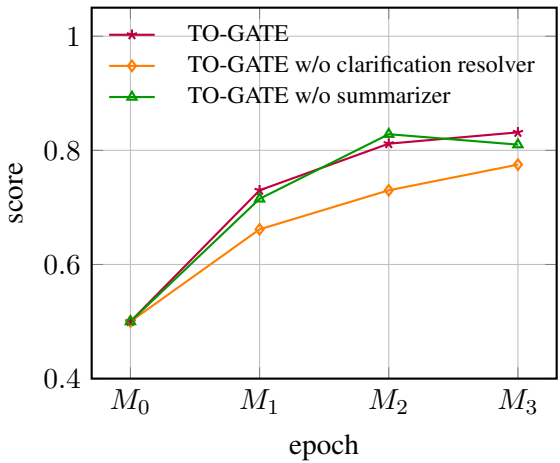

\subsection{Analysis}

\paragraph{Ablations}
We conducted an ablation study on TO-GATE, removing the clarification resolver and summarizer to assess their individual impact. As shown in Figure~\ref{fig:W/o weight}, the performance of TO-GATE drops significantly without these components. Specifically, the model without a clarification resolver experiences an 5.65\% reduction in win rate, while the model without a summarizer loses 2.12\% in win rate. These results highlight the critical roles of both modules in TO-GATE, enabling it to effectively elicit human preferences through clarifying questions and provide accurate responses to specific tasks. In particular, the clarification resolver has a more substantial impact on overall performance, suggesting that the quality of question-asking trajectories plays a more significant role in the model's success.
\vspace{-0.5em}
\paragraph{$\lambda$ in Response Loss}
We investigate $\lambda$ that is used to balance clarification and responses losses in Eq.~\ref{eq:weight}. As shown in Table~\ref{tab:weight adjuest}, the model with $\lambda = 2$ achieves the best performance compared to other settings, by obtaining 78.66\% win rate in the response evaluation. As $\lambda$ increases, the model overemphasizes the  responses and weakens the effect of clarifications, resulting in decreased discriminative ability and personalization performance. We conclude that the clarifying questions can elicit human preferences while enabling the questioner to summarize the response to the task is necessary.

\begin{table}[!tp]
\centering
\caption{Results of responses given by $M_1$ of TO-GATE with different $lambda$ values}
\label{tab:weight adjuest}
\begin{tabular}{lccc}
\toprule
$\lambda$ & \textbf{A-B} & \textbf{B-A} & \textbf{Average} \\
\midrule
1    & 71.08 & \textbf{84.35} & 77.72 \\
2 & \textbf{74.28} & 83.04 & \textbf{78.66} \\
3    & 73.28 & 81.16 & 77.20 \\
6    & 73.68 & 77.54 & 75.61\\
\bottomrule
\end{tabular}
\end{table}

\section{Conclusion}
To effectively elicit human preferences, we propose a novel training framework of TO-GATE with trajectory optimization, which consists of a clarification resolver that employs contrastive learning to penalize ineffective questions, and a summarizer that balances the quality of questions and responses. Additionally, we introduce the use of deterministic metrics to independently evaluate the model's performance on both clarifications and responses. Experimental results demonstrate that our model achieves state-of-the-art performance on standard human preference elicitation tasks.

\section*{Limitations}
The model performances of eliciting human preferences are limited to LLMs. The backends of the components in tasks (Oracle, Roleplayer and Questioner) are built on pre-trained LLMs. As the scale of the LLMs increases, the ability of eliciting human preference will increase accordingly.

The data used in the experiments is automatically generated by using GPT-4, aiming to simulate the open-domain conversations between agents and users. The limitation is that the real-world human preference elicitation could be slightly different from the simulations.

\bibliography{custom}

\clearpage
\appendix
\twocolumn

\section{Symbol Definitions}
\begin{table}[htbp]
\centering
\caption{Symbol Definitions} 
\begin{adjustbox}{max width=\linewidth}
\begin{tabular}{lp{0.75\linewidth}}
\toprule
\textbf{Symbol} & \textbf{Description} \\
\midrule
$\mathcal{T}$ & Task set \\
$\mathcal{U}$ & User set \\
${G}$   & Gold response set \\
$t_i$ & Task i in the task set \\
$u_j$ &  User j in the task set\\
$o^g_{ij}$ & Oracle-generated gold response for $t_i$ and $u_j$\\
$Q_{base}$ & Baseline model for evaluation\\
$Q$ & Questioner\\
$R$ & Reloplayer\\
$\mathcal{O}$   & Oracle (GPT-4) \\
$s_{ij}$   & The conversations for $t_i$ and ${u_j}$\\
$s^w_{ij}$   &  Filtered positive conversations\\
$s^l_{ij}$   &  Filtered negative conversations\\
$q_{ijk}$ & Questioner's question at turn k\\
$a_{ijk}$ & Reloplayer's answer at turn k\\
$q_w$      & Positive clarifying questions\\
$q_l$ &  Negative clarifying questions\\
$o_w$ & Positive final response \\
$o_l$ &  Negative final response\\
$\pi_0$ &  Initial policy model\\
$\pi_\theta$ &  Final policy model\\
$\pi_{ref}$ & Reference policy model\\
$M_0$ & Baseline model\\
$M_n$ &  Model after training iteration n \\
$\sigma$ &  Sigma function \\
$\beta$ &   Controlling KL divergence\\
$r$ &  Reward function \\
$Z$ & Partition function\\
$\mathcal{D}_p$ & The dynamic dataset\\
$\lambda$ & summarizer weight control parameter\\
$n$ & Number of training iterations\\
$\mathcal{L}_c$ & Clarification Resolver module loss \\
$\mathcal{L}_o$ & summarizer module loss \\
$\mathcal{L}$ &  Loss of To-GATE\\
\bottomrule
\end{tabular}
\label{tab:Symbol Definitions}
\end{adjustbox}
\end{table}

\section{Dataset Construction}
\label{sec:dataset-construction}

\subsection{Simulation Conversations}

\begin{figure}[htbp]
    \centering
    \includegraphics[width=0.9\linewidth]{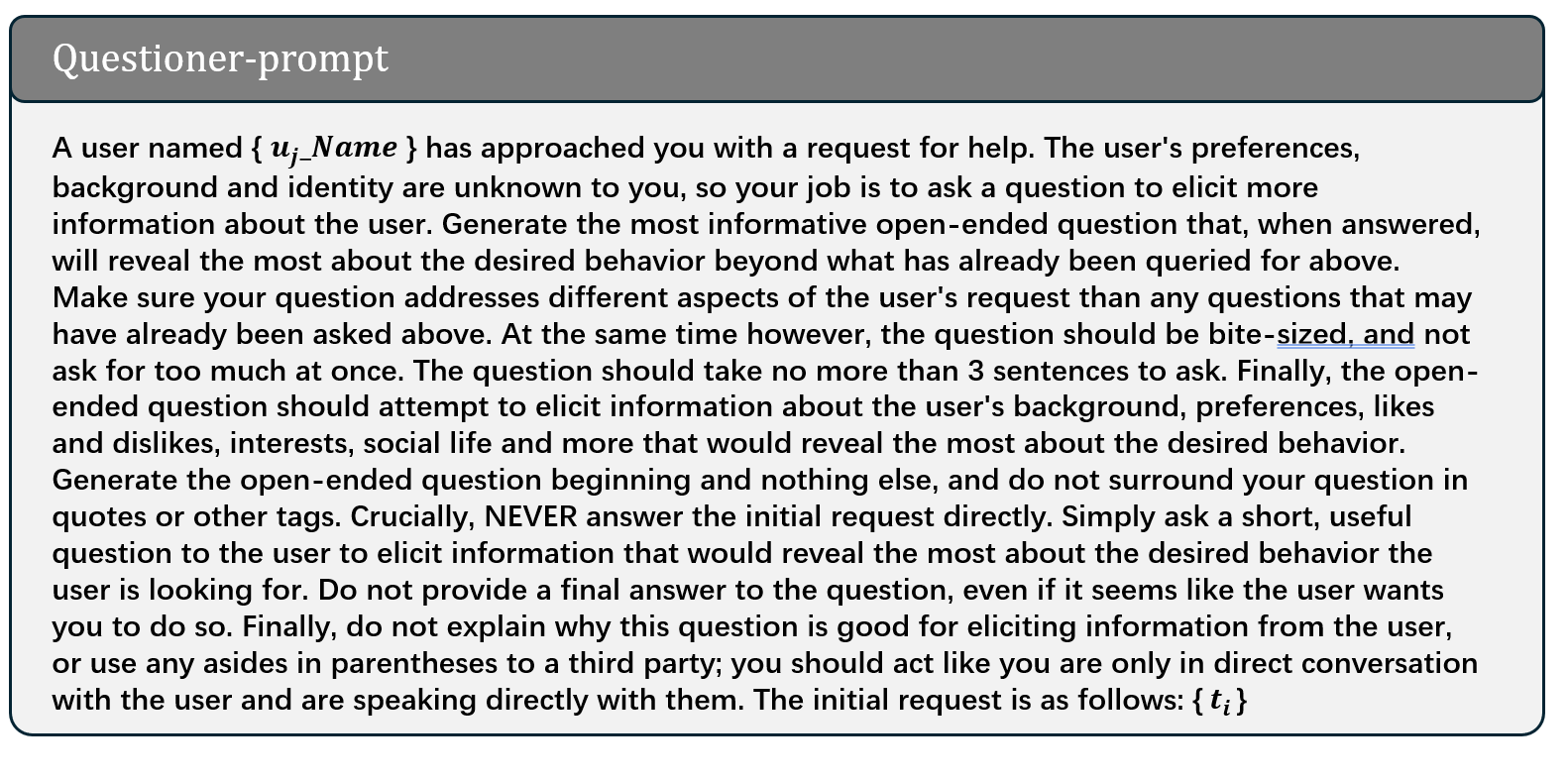}
    \caption{The prompt designed to guide the Questioner’s inquiries, providing direction and imposing certain constraints on the question format.}
    \label{fig:Questioner-prompt}
\end{figure}

\begin{figure}[htbp]
    \centering
    \includegraphics[width=0.9\linewidth]{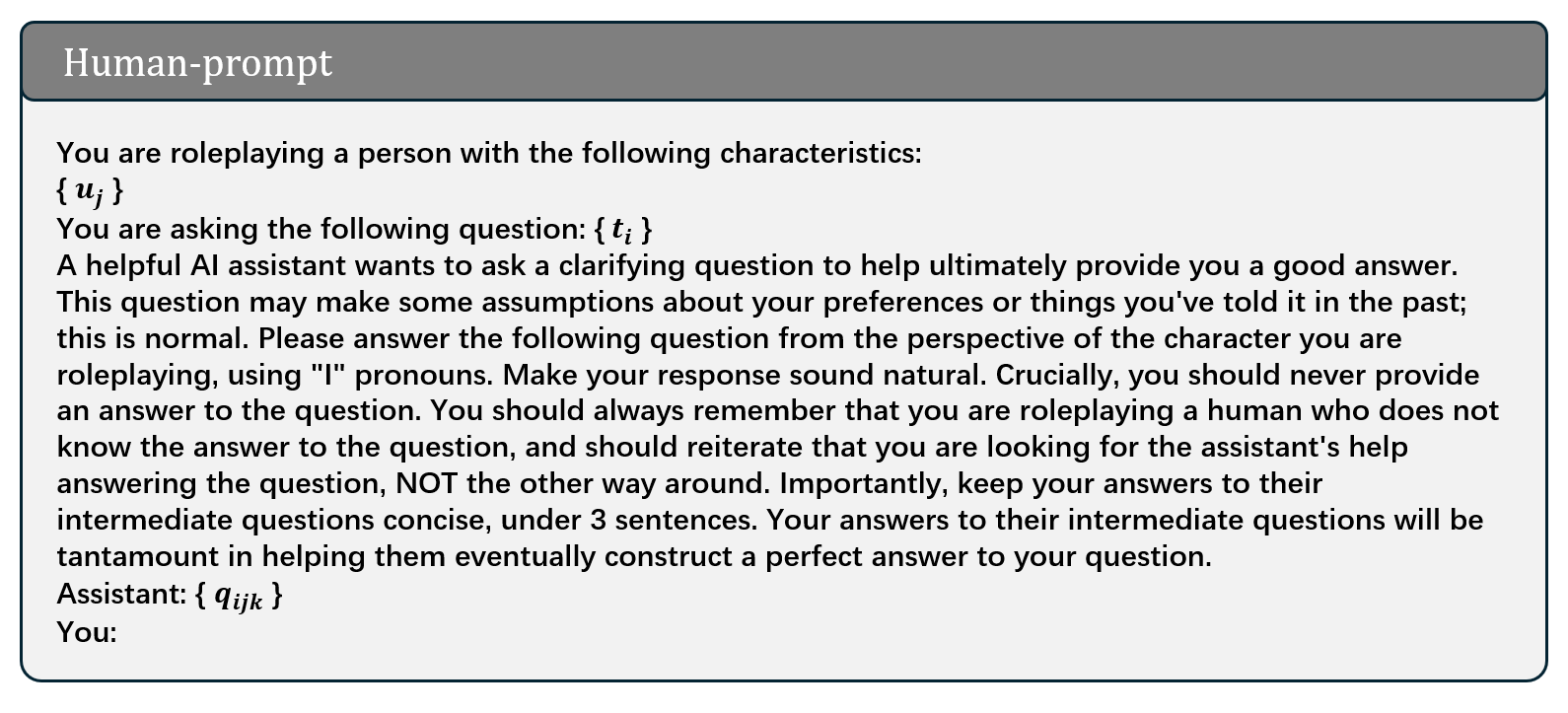}
    \caption{The prompt designed to guide the Roleplayer in responding to the Questioner's inquiries, with specific guidelines and constraints on the direction and format of the answers.}
    \label{fig:hunam-prompt}
\end{figure}

Our dataset consists of 110 user profiles and 550 task instructions. We use 100 user profiles and 500 task instructions as the training set, and 10 user profiles and 50 task instructions as the test set. We generate simulated Conversations by interacting each task $t_i$ with each character $u_j$. The simulated Conversations generation consists of two parts: the first part involves the Questioner asking questions to elicit task preferences (see~\ref{fig:Questioner-prompt}); the second part has the Roleplayer responding to the Questioner's questions (see Figure~\ref{fig:hunam-prompt}), forming one to three rounds of interactive dialogue.

\subsection{Gold Responses Construction}
\begin{figure}[htbp]
    \centering
    \includegraphics[width=0.9\linewidth]{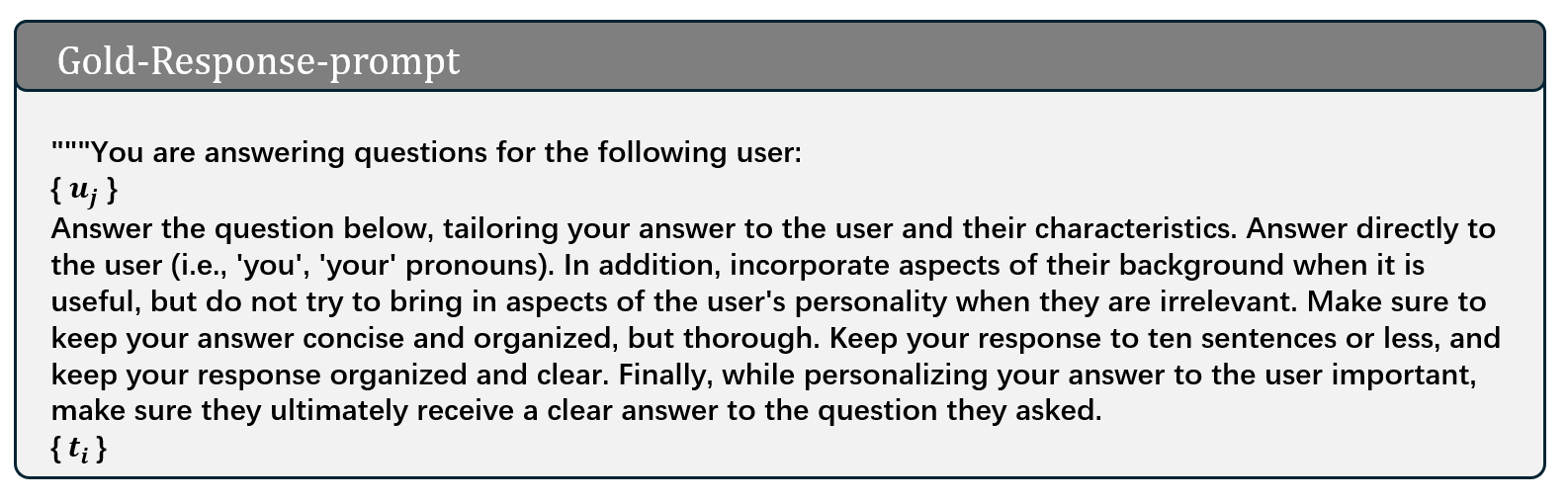}
    \caption{The prompt for generating personalized and precise high-quality responses by GPT-4}
    \label{fig:gold-response}
\end{figure}

Based on clear user background and task information, GPT-4 generates a personalized and precise high-quality response template. The first placeholder is used to insert detailed user background, and the second placeholder is for the initial query, ensuring the reply aligns with the user’s characteristics and adheres to the response format requirements,show in Figure~\ref{fig:gold-response}.

\end{document}